\documentclass[10pt,twocolumn,letterpaper]{article}

\usepackage{lmpaper}

\usepackage{cuted}
\usepackage{caption}
\usepackage{multirow}
\usepackage{array}
\usepackage{placeins}

\definecolor{linkblue}{rgb}{0.21,0.49,0.74}
\usepackage[pagebackref,breaklinks,colorlinks,allcolors=linkblue]{hyperref}

\title{Where Should the Lamp Go? Ambient-Aware Lamp Placement for Portraits}

\author{\href{https://yukyin.github.io/}{\textcolor{black}{Yuyan Chen}}\\
\href{https://modelslive.org/}{\textcolor{black}{ModelsLive Inc.}}\\
\href{mailto:ychen@modelslive.org}{\textcolor{black}{\tt\small ychen@modelslive.org}}
}

\begin{document}
\maketitle

\begin{strip}
\vspace{-32pt}
\centering
\includegraphics[width=\linewidth]{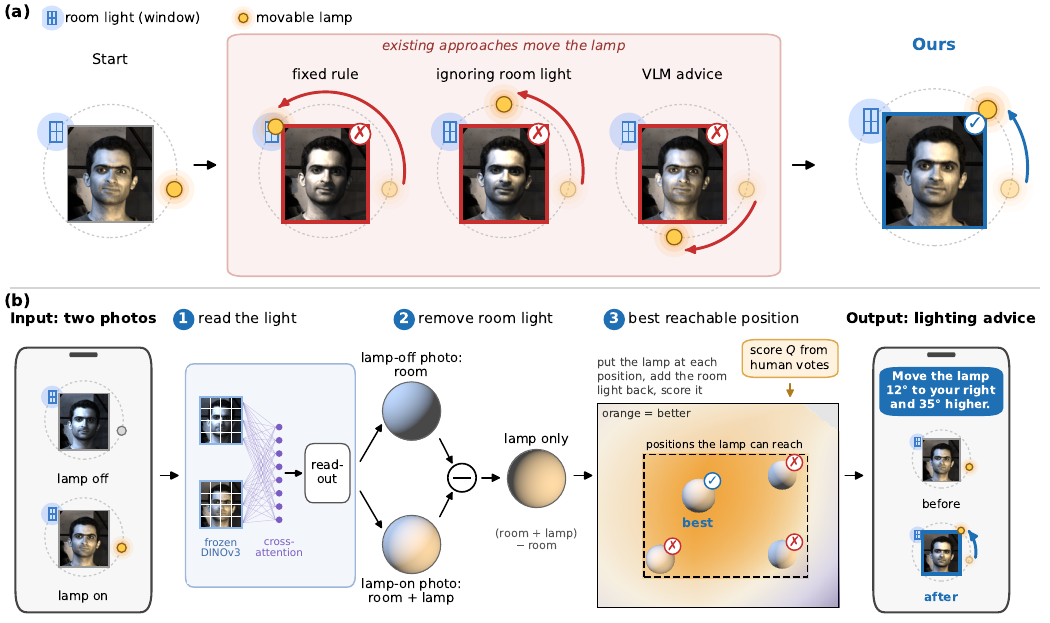}
\captionsetup{hypcap=false}
\captionof{figure}{\textbf{Where should the lamp go?} (a) A window already lights the face and one lamp can move. Each panel shows the lights around the face as seen from the camera and the resulting light on the face (window light blue, lamp light orange). From the same start, existing approaches move the lamp to a fixed position (fixed rule), overhead (ignoring the room light) or below the face (VLM advice). Ours accounts for the ambient light and better matches human aesthetic preference. (b) LightMatch. The input is two photos, lamp off and lamp on. In step 1, each photo is read by cross-attention over frozen DINOv3 features, and a read-out estimates the lighting. In step 2, subtracting the lamp-off estimate (room) from the lamp-on estimate (room + lamp) leaves the lamp alone. In step 3, the lamp is placed at every reachable position, the room light is added back, and each result is scored with $Q$, learned from human votes. The best position is returned as lighting advice, shown with the light before and after the move.}
\captionsetup{hypcap=true}
\label{fig:teaser}
\end{strip}

\begin{abstract}
Portraits shape first impressions, and lighting is an important factor. Prior work, such as lighting estimation and portrait relighting, edits lighting after capture but does not guide where to place a light before the shot. This leaves an open question: where should a light go to best light a face? Therefore, we propose LightMatch, a preference-grounded inverse lighting framework that disentangles the controllable light from ambient illumination and optimizes its placement. We first learn a Bradley-Terry quality model from pairwise human judgments. A perception head then regresses spherical-harmonic lighting from frozen self-supervised features via cross-attention. Finally, a placement solver isolates the light via spherical-harmonic linearity and optimizes its reachable position. On held-out environments, LightMatch attains 87.8\% of the oracle improvement, outperforming the conventional solver that ignores ambient light by a relative 28.4\% and the same pipeline with the widely used DPR estimator by a relative 70.5\%. It reaches 73.2\% on real OLAT faces of unseen identities and, on our real captures, cuts lamp-direction error by 37.7\%.
\end{abstract}

\section{Introduction}
\label{sec:intro}

A portrait is often the first impression a person makes, and lighting largely decides how a face reads. Portraits are now taken everywhere, from profile pictures and video interviews to live streams and product videos. Yet the ambient light is often not enough, so one or more lamps are added. The question is where the lamps should go so that the combined lighting flatters the face more. For example, in \cref{fig:teaser}(a), light from a window falls on the face as part of the ambient light. The classic Rembrandt rule puts the lamp at a fixed position regardless of the window, making the face look harsher, while a lamp placed with the window light taken into account makes the face look more even, which is the better position we aim to find.

Existing work addresses neighbouring problems. \emph{Portrait relighting}~\cite{kim2024switchlight,pandey2021total,sun2019single,zhou2019deep} edits pixels after capture toward a target light chosen by the user, but does not change the real scene. \emph{Lighting estimation}~\cite{gardner2017learning,hold2017deep,legendre2019deeplight,phongthawee2024diffusionlight} and \emph{face-model fitting}~\cite{feng2021learning} recover the light in a scene, but have no notion of which light is better. \emph{Aesthetics models}~\cite{murray2012ava,talebi2018nima} predict how good a whole photo looks, but cannot tell which change to the lighting would improve it. \emph{Vision-language models} (VLMs)~\cite{liu2023visual} can suggest such changes in words, but their visual perception still lags behind humans, by about 15 percentage points on even Gemini 2.5 Pro, one of the leading models~\cite{kamoi2024visonlyqa}.

We formalize this as \emph{ambient-aware lamp placement}, a constrained inverse lighting problem: given observations of the scene, find the lamp pose within its reachable range that maximizes the perceived quality of the combined lighting on the face. We study its simplest form, with a single movable lamp observed in one lamp-off and one lamp-on photo. To solve it, we introduce LightMatch, a preference-grounded inverse lighting framework with three components, shown in \cref{fig:teaser}(b). A quality model, the Bradley--Terry score $Q$~\cite{bradley1952rank} over light direction and key-to-fill ratio, is fitted to pairwise human judgments. A perception head reads spherical-harmonic (SH) lighting from each photo by cross-attending to frozen DINOv3~\cite{simeoni2025dinov3} features, trained on renders and real OLAT composites. A placement solver isolates the lamp by differencing the two SH readings and searches the movable range for the position that maximizes $Q$ under the ambient light.

Across held-out environments, LightMatch attains 87.8\% of the oracle improvement, against 68.4\% for a solver that ignores the ambient light, and it still recovers 73.2\% on real faces of unseen identities. The quality score $Q$ agrees with professional photographers with a correlation of 0.86.

Our contributions are as follows:
\begin{itemize}[leftmargin=*,nosep]
\item To our knowledge, we are the first to pose ambient-aware lamp placement from real photographs, a preference-driven inverse lighting problem that couples physical light perception with human aesthetics.
\item We develop LightMatch, which uses frozen self-supervised features, disentangles the controllable light through spherical-harmonic linearity, and places it by learned preference.
\item LightMatch beats an ambient-unaware solver by a relative 28.4\%, and its quality model correlates at 0.86 with one refitted on professional photographers' votes.
\end{itemize}

\section{Related Work}
\label{sec:related}

\paragraph{Relighting and lighting design.} Pandey et al.~\cite{pandey2021total} and Kim et al.~\cite{kim2024switchlight} predict normals and albedo to relight portraits for background replacement. Chaturvedi et al.~\cite{chaturvedi2025synthlight} and Zhang et al.~\cite{zhang2025scaling} relight with diffusion models, trained to re-render synthetic faces and constrained by consistent light transport respectively. Schoeneman et al.~\cite{schoeneman1993painting} and Pellacini et al.~\cite{pellacini2007lighting} solve for light parameters that reproduce a painted target in a rendered scene. Srikanth et al.~\cite{srikanth2014computational} move a light on an aerial robot to achieve rim lighting from camera feedback, and Jiang and Chen~\cite{jiang2026before} plan lights, camera and pose for portraits in a full 3D scene. These methods edit pixels after capture, require a complete 3D scene or adjust the light by physical trial and error, leaving open where to place a physical light from photographs alone, which motivates our placement solver.

\paragraph{Lighting estimation and visual perception.} Phongthawee et al.~\cite{phongthawee2024diffusionlight} and Chinchuthakun et al.~\cite{chinchuthakun2026diffusionlight} estimate scene lighting by inpainting a chrome ball with a diffusion model, and Liang et al.~\cite{liang2026luxdit} generate HDR environment maps with a video diffusion transformer. Feng et al.~\cite{feng2021learning} fit a FLAME face model and recover SH lighting as a by-product. Kamoi et al.~\cite{kamoi2024visonlyqa} and Dahou et al.~\cite{dahou2025vision} show that VLMs trail humans on simple geometric and saliency perception. El Banani et al.~\cite{el2024probing} find that self-supervised backbones encode 3D structure best. These methods perceive the light that is present but cannot isolate a movable lamp from it, a gap that motivates our perception head and differencing.

\paragraph{Aesthetics and preference learning.} Murray et al.~\cite{murray2012ava} collect large-scale aesthetic ratings, and Cao et al.~\cite{cao2026artimuse} assess aesthetics with a multimodal LLM that gives both scores and expert analysis. Bradley and Terry~\cite{bradley1952rank} turn pairwise comparisons into scores. Xu et al.~\cite{xu2023imagereward} and Ma et al.~\cite{ma2025hpsv3} learn preference reward models for generated images. These models score a finished image but do not reveal how a lighting change would affect it, which inspires our quality model over lighting parameters.

\section{Judging Portrait Lighting}
\label{sec:q}

\paragraph{Definition.} We define good portrait lighting by human preference: one lighting is better than another if people choose it when shown both. A quality score $Q$ captures this preference as a function of three parameters of the key light: azimuth $\alpha$, positive toward the subject's right, elevation $\epsilon$, and key-to-fill ratio $\rho$, the magnitude of the first-order SH term relative to the uniform term (\cref{eq:v}).

\begin{figure}[t]
\centering
\includegraphics[width=\linewidth]{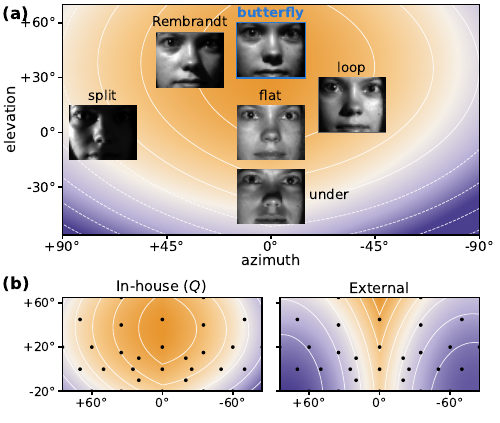}
\caption{\textbf{What $Q$ prefers.} (a) $Q$ over lamp position at $\rho = 2$, with purple to orange representing low to high $Q$. Each face is a real photograph of a classic setup, placed at its lamp position, and the best of the six, butterfly lighting, is marked by the blue frame. (b) $Q$ from in-house votes and from the votes of four external photographers, at the 30 lamp positions of the Yale faces (dots).}
\label{fig:landscape}
\end{figure}

\paragraph{Modeling.} Existing scorers fail at this task, as CLIP~\cite{radford2021learning} ranks flat or split light first among six classic setups and responds more to the face than to the light. To build $Q$, we judged pairs of rendered FLAME heads~\cite{li2017learning} with varied identity and makeup, starting from a round of 80 pairs chosen by experience and discarding pairs that could not be told apart. Later rounds widened the elevation range until the fitted optimum lay inside it, leaving 382 votes. With $a=\alpha/90^\circ$ and $e=\epsilon/60^\circ$ scaling the two angles to a similar range, we fit a quadratic
\begin{equation}
Q(\alpha,\epsilon,\rho)=\mathbf{w}^\top[a,e,\rho,a^2,e^2,\rho^2,ae,a\rho,e\rho,|a|],
\label{eq:q}
\end{equation}
with weights $\mathbf{w}$ estimated by Bradley--Terry logistic regression. The linear and squared terms give $Q$ a single best lighting, the cross terms let direction and ratio interact, and the $|a|$ term makes the preference fall off as the light moves away from the front. This is the simplest such form, and its ten weights are reliably estimated from a few hundred votes. Interaction terms with identity, makeup and bone structure did not improve held-out accuracy, with all $p\geq0.30$, so one shared $Q$ is used across faces. $Q$ peaks at azimuth $0^\circ$, elevation $+35^\circ$ and ratio $2.0$. With a single lamp, this best position lies next to butterfly lighting, a standard setup in portrait photography~\cite{grey2014master}. $Q$ also ranks six classic setups in the textbook order~\cite{grey2014master}, with $Q=2.19$ for butterfly, 1.54 for loop, 1.27 for Rembrandt, 1.19 for flat, $-1.31$ for split and $-1.48$ for under-light at ratio 1. \Cref{fig:landscape}(a) places these setups on the map of $Q$. This agreement with established photographic practice supports $Q$ as a measure of lighting quality. It predicts 79.3\% of held-out votes, indicating that it can serve as a reasonably reliable proxy for portrait lighting quality in this work.

\paragraph{Alignment with photographers.} To check whether $Q$ reflects professional taste beyond our own judgments, we recruited six professional photographers, who volunteered without payment, with 3 to more than 10 years of experience in portrait, wedding and landscape photography. We randomly sampled 200 pairs for each photographer from real Yale faces~\cite{georghiades2001few,lee2005acquiring} of 27 subjects under 30 light positions and three fill ratios, with face brightness equalized, and asked them to choose the better-lit image in each pair. One pair in ten repeated an earlier pair with sides swapped, so that we could measure how consistent each photographer was. Only photographers who made the same choice on at least 80\% of the repeats were kept, so we discarded the votes of two of the six. In leave-one-out evaluation, a Bradley--Terry model fitted to the votes of three of the remaining four predicts 61.1\% to 77.2\% of the choices of the fourth, above the 50\% chance level, showing that the four judge lighting by the same standard. \Cref{fig:landscape}(b) compares $Q$ obtained from in-house votes with $Q$ obtained from the votes of the four photographers, and the two have a correlation of 0.86 over the 30 lamp positions. $Q$ also agrees with 64.7\% of the photographers' choices, close to the 72.8\% reached by a $Q$ fitted on their own votes, indicating that $Q$ is consistent with the judgment of professional photographers. More details are in \cref{sec:s-photo-study}.

\section{Reading the Light}
\label{sec:read}

This step reads the light on the face from a photograph in order to obtain the lighting parameters that $Q$ scores, step 1 in \cref{fig:teaser}(b). The light in a real room arrives from many directions at once, so we describe it by a second-order SH expansion. The expansion contains nine coefficients, which together represent the light from all directions, and it matches the diffuse reflection of skin~\cite{ramamoorthi2001efficient}:
\begin{equation}
L(\boldsymbol{\omega})\approx\sum_{i=0}^{8}c_i\,Y_i(\boldsymbol{\omega}),
\label{eq:sh}
\end{equation}
where $Y_i$ are the SH basis functions and $\mathbf{c}=(c_0,\dots,c_8)$ are the coefficients to be read, with $c_0$ the uniform term and $\mathbf{c}_{1:3}$ the first-order term, which points toward the dominant light.

\begin{figure}[t]
\centering
\includegraphics[width=\linewidth]{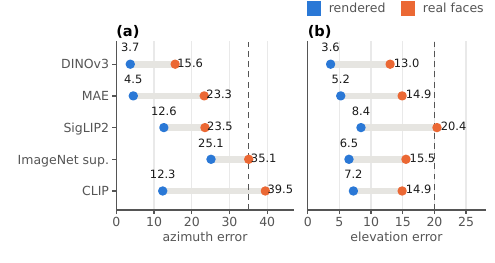}
\caption{\textbf{Performance and transfer of different vision backbones.} (a) Azimuth error. (b) Elevation error. Heads are trained on 2,000 additional renders and tested on 500 held-out renders and on real Yale faces, shown as blue and orange dots, and a longer bar means a weaker transfer from renders to real faces. For both blue and orange dots, further left is better. The dashed line marks the error of guessing the same direction for every image without looking at it, and a point left of the line, that is, with a smaller error, shows that the backbone indeed reads the lighting from the image.}
\label{fig:backbones}
\end{figure}

Exact light labels are available in large numbers only for rendered faces, yet the reader must work on real photos, which requires image features that transfer from renders to real faces. A natural candidate is a large VLM, which has seen many real images. In a preliminary study, however, it failed to transfer, with 40 to 67$^\circ$ azimuth error on real faces across its layers, worse than the 35$^\circ$ of guessing (\cref{sec:ablation}). Among five frozen backbones, DINOv3 transferred best (\cref{fig:backbones}), so we build a perception head on DINOv3~\cite{simeoni2025dinov3}. To preserve this transfer, we keep the backbone frozen, as fine-tuning it on renders would adapt its features to rendered appearance. The head is trained on two kinds of data: rendered faces, which can provide exact labels, and composites of real one-light-at-a-time captures, which can provide real skin appearance. We also use mirror augmentation to keep left and right consistent.

Specifically, the frozen backbone first encodes $I$ into patch features $Z=\phi(I)$, one per face region, and a linear layer $W$ projects them to width $d$. Eight learned queries $U^{(0)}\in\mathbb{R}^{8\times d}$ then collect the shading spread over the cheeks, nose and brow through two blocks,
\begin{equation}
U^{(l+1)}=\mathrm{Block}\big(U^{(l)},\,WZ\big),\quad l=0,1,
\label{eq:block}
\end{equation}
each consisting of cross-attention from the queries to the patches, self-attention among the queries and an MLP. Finally, the eight queries are averaged and an MLP regresses the nine coefficients,
\begin{equation}
\hat{\mathbf{c}}=\mathrm{MLP}\Big(\tfrac{1}{8}\textstyle\sum_{j=1}^{8}U^{(2)}_j\Big).
\label{eq:head}
\end{equation}
Since $Q$ takes the key-light direction and ratio rather than SH coefficients, we convert $\hat{\mathbf{c}}$ into them. The first-order term points toward the dominant light, and its size relative to the uniform term measures how much stronger the key is than the fill, so
\begin{equation}
\mathbf{v}=\hat{\mathbf{c}}_{1:3}/\hat{c}_0,\qquad \rho=\|\mathbf{v}\|,
\label{eq:v}
\end{equation}
where $\alpha$ and $\epsilon$ are the direction of $\mathbf{v}$ and $\rho$ is its length.

\section{Placing the Lamp}
\label{sec:ambient}

This step finds where to place the lamp so that the light on the face is best, steps 2 and 3 in \cref{fig:teaser}(b). The light on a face usually also includes ambient light from the room. To know what the movable lamp alone contributes, we compare a lamp-off photo $I_{\mathrm{off}}$ and a lamp-on photo $I_{\mathrm{on}}$ taken at a locked exposure, similar to flash and no-flash pairs~\cite{aksoy2018dataset}. The lamp's own SH coefficients describe its current direction and strength on the face, from which we can compute a better position. Since light adds, these coefficients are the difference between the readings of the two photos by the perception head $h$,
\begin{equation}
\hat{\mathbf{c}}_a=h(I_{\mathrm{off}}),\qquad \hat{\mathbf{c}}_k=h(I_{\mathrm{on}})-h(I_{\mathrm{off}}).
\label{eq:diff}
\end{equation}
For a single lamp, the first-order term is a vector that points toward the lamp and whose length gives its strength, $\hat{\mathbf{c}}_{k,1:3}=s_k\mathbf{d}_k$, which yields the current lamp direction $\mathbf{d}_k$ and strength $s_k$.

To find a better position, we move the lamp to a candidate direction $\mathbf{d}$ with the same strength and add the ambient light back,
\begin{equation}
\hat{\mathbf{c}}_{1:3}(\mathbf{d})=\hat{\mathbf{c}}_{a,1:3}+s_k\mathbf{d},\qquad \hat{c}_0=\hat{c}_{a,0}+\hat{c}_{k,0},
\label{eq:move}
\end{equation}
and score the combined light with $Q$ at the direction and ratio given by \cref{eq:v}. In this way, the $Q$ learned from human votes scores the lighting of any room without new annotation, and the ambient light decides which lamp position is best. We then select the reachable direction with the highest score,
\begin{equation}
\mathbf{d}^\ast=\arg\max_{\mathbf{d}\in\mathcal{R}}\;Q\big(\alpha(\mathbf{d}),\epsilon(\mathbf{d}),\rho(\mathbf{d})\big),
\label{eq:solve}
\end{equation}
where $\mathcal{R}$ is the set of directions the lamp can reach, searched on a $5^\circ$ grid. $\mathcal{R}$ keeps only positions whose combined lighting stays within the range covered by the human votes (azimuth $\pm90^\circ$, elevation $-30^\circ$ to $75^\circ$, ratio 0.15 to 3.8), where $Q$ is fitted. Finally, a fixed template turns the change in azimuth and elevation from the current direction $\mathbf{d}_k$ to $\mathbf{d}^\ast$ into lighting advice in words, such as moving the lamp $12^\circ$ to the subject's right and $35^\circ$ higher, which is the output in \cref{fig:teaser}(b). 

\begin{table*}[t]
\centering
\small
\setlength{\tabcolsep}{3.5pt}
\caption{Lamp placement on the four test sets. For the fourth test set, only the lamp direction is known, so it is evaluated by Error and by Rating.}
\label{tab:main}
\resizebox{\textwidth}{!}{%
\begin{tabular}{@{}lcccccccccccccc@{}}
\toprule
& \multicolumn{4}{c}{Standard} & \multicolumn{4}{c}{Real ambient light} & \multicolumn{4}{c}{Real faces} & \multicolumn{2}{c}{Real capture} \\
\cmidrule(lr){2-5}\cmidrule(lr){6-9}\cmidrule(lr){10-13}\cmidrule(l){14-15}
Method & Share$\uparrow$ & Worse$\downarrow$ & Recover$\uparrow$ & Error$\downarrow$ & Share$\uparrow$ & Worse$\downarrow$ & Recover$\uparrow$ & Error$\downarrow$ & Share$\uparrow$ & Worse$\downarrow$ & Recover$\uparrow$ & Error$\downarrow$ & Error$\downarrow$ & Rating$\uparrow$ \\
\midrule
DECA & 4.1 & 43.9 & 37.5 & 68.4 & 2.6 & 45.0 & 10.5 & 69.4 & 7.3 & 41.8 & 22.5 & 52.2 & 101.3 & -0.7 \\
Fixed Rembrandt & 22.4 & 34.0 & 70.0 & -- & 25.1 & 32.1 & 26.3 & -- & 25.9 & 31.4 & 42.5 & -- & -- & 1.1 \\
DPR & 51.5 & 12.5 & 55.0 & 42.4 & 52.9 & 11.3 & 31.6 & 45.0 & 36.3 & 24.2 & 50.0 & 48.7 & 50.4 & 1.8 \\
Ambient-unaware & 68.4 & 9.6 & 67.5 & 8.5 & 71.0 & 9.4 & 42.1 & 7.6 & 62.4 & 9.2 & 57.5 & 11.5 & 28.9 & 2.0 \\
LightMatch & \textbf{87.8} & \textbf{2.6} & \textbf{75.0} & \textbf{5.7} & \textbf{89.1} & \textbf{2.2} & \textbf{57.9} & \textbf{5.8} & \textbf{73.2} & \textbf{8.6} & \textbf{72.5} & \textbf{8.2} & \textbf{18.0} & \textbf{2.4} \\
\bottomrule
\end{tabular}%
}
\end{table*}

\begin{figure}[t]
\centering
\includegraphics[width=\linewidth]{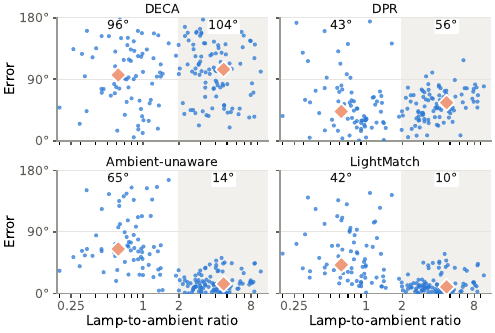}
\caption{\textbf{Lamp-direction error on the fourth test set against the lamp-to-ambient ratio.} Each point is one photo. The shaded area marks a lamp twice as strong as the ambient light or stronger. The diamond on each side marks the median error, and the number at the top gives its value.}
\label{fig:capture}
\end{figure}

\begin{figure*}[t]
\centering
\includegraphics[width=\textwidth]{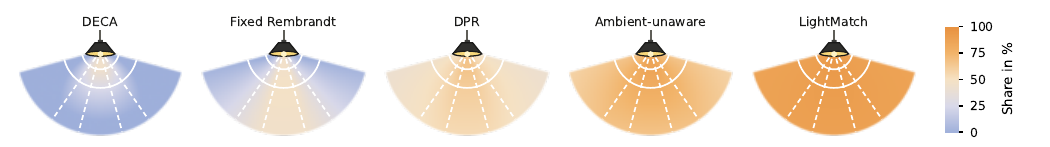}
\caption{\textbf{Share under different movable ranges of the lamp, averaged over the first two test sets.} Each fan shows the range of azimuth and elevation within which the lamp can move. Colors at azimuth $\pm15^\circ$, $\pm35^\circ$ and $\pm75^\circ$, marked by the white dashed lines and the straight sides of the fan, and at elevation $\pm10^\circ$, $\pm20^\circ$ and $\pm45^\circ$, marked by the white solid lines and the outer arc of the fan, are measured, and colors in between are interpolated.}
\label{fig:range}
\end{figure*}

\begin{table}[t]
\centering
\small
\setlength{\tabcolsep}{3pt}
\caption{Ablation on the Standard, Real ambient light and Real faces test sets.}
\label{tab:ablation}
\resizebox{\columnwidth}{!}{%
\begin{tabular}{@{}llcccc@{}}
\toprule
Test set & Variant & Share & Worse & Recover & Error \\
\midrule
\multirow{2}{*}{Standard} & LightMatch & \textbf{87.8} & \textbf{2.6} & \textbf{75.0} & \textbf{5.7} \\
 & w/o lamp-off & 80.4 & 3.8 & 67.5 & 7.3 \\
\midrule
\multirow{2}{*}{Real ambient light} & LightMatch & \textbf{89.1} & \textbf{2.2} & \textbf{57.9} & \textbf{5.8} \\
 & w/o lamp-off & 82.3 & 2.4 & 52.6 & 6.6 \\
\midrule
\multirow{3}{*}{Real faces} & LightMatch & \textbf{73.2} & \textbf{8.6} & \textbf{72.5} & \textbf{8.2} \\
 & w/o lamp-off & 67.5 & 9.8 & 57.5 & 10.9 \\
 & w/o real-face & 30.6 & 29.1 & -- & 34.5 \\
\bottomrule
\end{tabular}%
}
\end{table}

\begin{table*}[t]
\centering
\small
\setlength{\tabcolsep}{4pt}
\caption{Lamp-placement advice on the no-ambient variant of the first test set. The variant has 800 renders generated in the same way without ambient light. The columns from GPT-6.1 Sol to Qwen3.5-27B are VLMs that give advice directly. Every model writes advice with the same prompt and parser.}
\label{tab:advice}
\resizebox{\textwidth}{!}{%
\begin{tabular}{@{}lccccccccc@{}}
\toprule
& LightMatch & w/ VLM read-out & w/ VLM advice & w/o read-out & GPT-6.1 Sol & Gemini 3.1 Pro & Claude Fable 5.1 & Grok 4.7 & Qwen3.5-27B \\
\midrule
Share$\uparrow$ & \textbf{96.2} & 94.8 & 94.0 & 90.5 & 22.6 & 25.2 & 7.5 & 11.4 & 0.8 \\
Worse$\downarrow$ & \textbf{1.2} & 1.4 & 1.4 & 3.1 & 31.7 & 29.7 & 37.7 & 32.3 & 31.5 \\
\bottomrule
\end{tabular}%
}
\end{table*}

\begin{figure}[t]
\centering
\includegraphics[width=\linewidth]{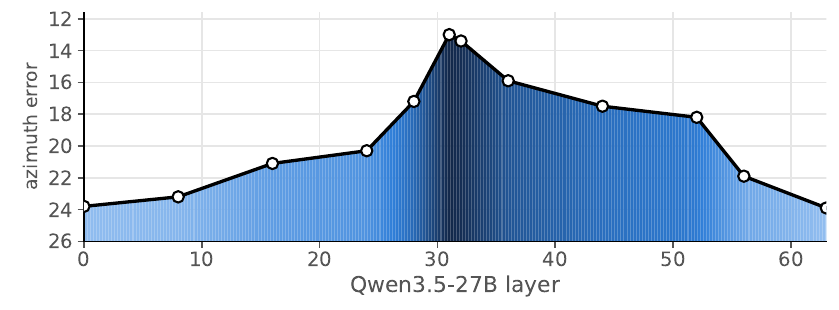}
\caption{\textbf{Decodability of lighting across the layers of the VLM.} Azimuth error of a linear probe on each layer of Qwen3.5-27B, trained and tested on renders. Higher points mean lower error, and layers 31 to 32 read the light best.}
\label{fig:probe}
\end{figure}

\begin{figure*}[t]
\centering
\includegraphics[width=\textwidth]{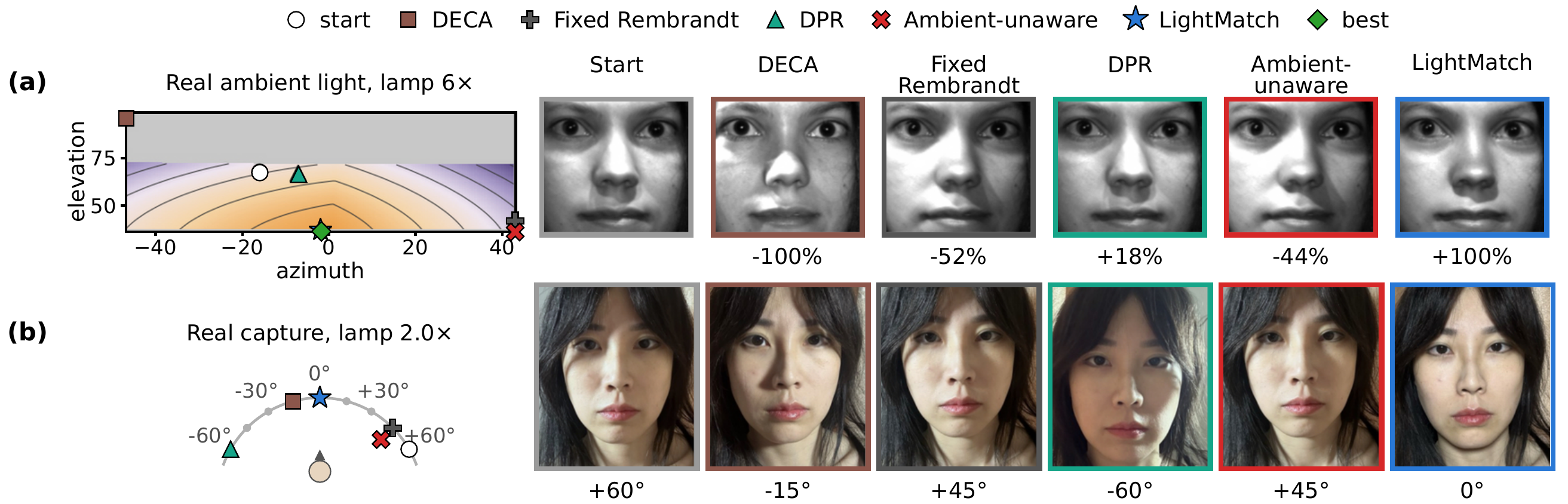}
\caption{Two examples with real ambient light. (a) An example from the Real ambient light test set with the lamp 6 times the ambient light, shown on a real Yale face. The left map shows $Q$ of the combined light at each lamp position within the movable range, with purple to orange representing low to high $Q$, gray outside the range covered by human evaluation, and the Share of each method below its face. (b) A real photo from the Real capture test set with the lamp 2 times the ambient light. The leftmost diagram shows the nine lamp positions seen from overhead, and the lamp position is below each photo. The markers show the start, the position chosen by each method, and the best position.}
\label{fig:examples}
\end{figure*}

\section{Experiments}
\subsection{Setup}
\label{sec:setup}

\paragraph{Implementation details.} The perception head uses a frozen DINOv3 ViT-B backbone. Each photo is padded to a square and resized to 336\,px. The head has eight queries of width 512 and two blocks with eight attention heads each. As the backbone is frozen, its features are computed once and only the head is trained. The head is first trained on the rendered pairs for 20 epochs at a learning rate of $3\times10^{-4}$, selecting the epoch by validation loss, on one A100 GPU. It is then trained further on 2,000 of the rendered pairs and the real-face pairs for 4 epochs at a learning rate of $10^{-4}$, which runs on a CPU. Training details of the ablation variants are in \cref{sec:s-yale}.

\paragraph{Data.} Our training data consist of lamp-off and lamp-on pairs in two parts. The first part has 4,000 rendered pairs. We create 4,000 FLAME faces~\cite{li2017learning} and select 270 Poly Haven panoramas~\cite{haven2025poly} as ambient light. The lamp direction is drawn at random from azimuth $\pm90^\circ$ and elevation $-30^\circ$ to $75^\circ$. The second part has 1,500 real-face pairs, 300 for each of five POLAR subjects, composed from the 149 POLAR lights with the same 270 panoramas. In both parts, faces, ambient light and lamps are combined at random. We test on four sets. The first has 400 rendered pairs with ambient light from the remaining 30 Poly Haven panoramas. The second has 400 rendered pairs with ambient light from the 20 measured rooms of SPLiT~\cite{fei2023split} and tests real ambient light. The third has 400 real-face pairs from the remaining five POLAR subjects and tests real faces. The fourth has 189 pairs of real photos of one subject, taken by us in one room under seven ambient conditions and 27 lamp positions, as an out-of-distribution test. The photos are taken with a phone on a tripod with exposure and focus locked. A floor lamp is placed at nine azimuths from $-60^\circ$ to $+60^\circ$ on three rings at elevations of $40^\circ$, $25^\circ$ and $12^\circ$, with lamp-off photos taken under each ambient condition. All participants, including the photographers in \cref{sec:q}, volunteered and gave informed consent, and the subject agreed to publish the photos. No IRB review was carried out, as the study only takes photos under household lamps and collects image choices. More details are in \cref{sec:s-ambient}.

\paragraph{Baselines.} We compare with five lamp placement methods. \emph{Fixed Rembrandt} always places the lamp at the classic Rembrandt position~\cite{hunter2021light}, at azimuth $45^\circ$ and elevation $40^\circ$, representing practice based on experience alone. \emph{Ambient-unaware} counts all ambient light as lamp light, as if the lamp were the only light source, and tests whether modeling the ambient light is necessary. \emph{DPR} and \emph{DECA} read the light with a portrait relighting network~\cite{zhou2019deep} and a 3D face reconstruction model~\cite{feng2021learning}, representing existing lighting estimators. Their outputs are calibrated to our lighting representation on renders, which favours them, and Fixed Rembrandt is given the true lamp position. DiffusionLight~\cite{phongthawee2024diffusionlight,chinchuthakun2026diffusionlight} is not compared, as on portraits it inpaints the chrome ball over the face and its estimate follows the background. \emph{Oracle} knows the true ambient and lamp light and gives the upper bound.

\paragraph{Metrics.} The movable range of the lamp is random in each test pair. \emph{Share}, the main metric, is the gain in $Q$ achieved by a method divided by the largest gain reachable within the range, so 100\% means the best position is found. \emph{Worse} is the fraction of pairs in which $Q$ drops after the move. \emph{Recover} is the fraction of pairs starting outside the range covered by human votes that are brought back inside. \emph{Error} is the median angle between the estimated and true lamp directions. Lower is better for Worse and Error, and higher for the others. Share, Worse and Recover are in \%, and Error is in degrees.

\subsection{Main Results}
\label{sec:main}

This section compares LightMatch with the baselines in \cref{tab:main}.
On the first test set (labeled as Standard), LightMatch beats Ambient-unaware, the strongest baseline, by 28.4\% in Share, 32.9\% in Error and 72.9\% in Worse, so it places the lamp best, reads it most accurately and worsens the lighting least often. Fixed Rembrandt trails LightMatch by only 6.7\% in Recover but by 74.5\% in Share, so a fixed lamp fixes unacceptable lighting but stays far from the best position. DPR and DECA trail by 41.3\% and 95.3\% in Share, so lighting estimates can hardly guide lamp placement. With Share computed by the $Q$ refitted on the photographers' votes, LightMatch still leads on all three test sets (\cref{tab:pro_q}). Moving the lamp toward the single-lamp optimum of $Q$ reaches only 34.3\% in Share on this test set (\cref{tab:single_opt}), as the best position moves with the ambient light, and on the fourth test set LightMatch recommends $0^\circ$ in only 8.5\% of the pairs (\cref{tab:azimuth_dist}).

On the second test set (labeled as Real ambient light), LightMatch stays within 1.5\% in Share and 1.8\% in Error of the first test set, beats Ambient-unaware by 25.5\% in Share, 23.7\% in Error and 76.6\% in Worse, and leads in 19 of 20 rooms, so separating the ambient light still works.

On the third test set (labeled as Real faces), LightMatch rises 43.9\% in Error and drops 16.6\% in Share from rendered faces, and DPR drops 29.5\% in Share, far more than under real ambient light, so faces matter more than ambient light. 

On the fourth test set (labeled as Real capture), LightMatch has a 37.7\% lower Error than Ambient-unaware, above the 32.9\% on the first test set, so separating the ambient light helps even more on real photos. In \cref{fig:capture}, the Error is 37.1$^\circ$ to 68.2$^\circ$ below twice the ambient light and 6.1$^\circ$ to 15.3$^\circ$ above it, with the sign of the lamp azimuth always correct, a clear boundary at about twice the ambient light. The first two test sets show the same trend, as shown in \cref{tab:breakdown}, where we group the pairs by lamp strength into weak, medium and strong, with the lamp below 2 times, 2 to 5 times and at least 5 times the ambient light, the lower threshold matching the boundary in \cref{fig:capture}. With a strong lamp, LightMatch reaches 92.1\% and 89.9\% in Share. With a weak lamp, it drops to 70.0\% and 84.4\%, which indicates that its advice becomes less accurate as the lamp weakens, while it still leads the other methods.

\begin{table}[t]
\centering
\small
\caption{Share in percent on the first two test sets, split by the strength of the lamp relative to the ambient light. Error is the lamp-direction error of LightMatch in degrees.}
\setlength{\tabcolsep}{3.5pt}
\begin{tabular}{@{}lrcccc@{}}
\toprule
 & & w/o & Ambient- & Fixed & \\
Lamp & LightMatch & lamp-off & unaware & Rembrandt & Error \\
\midrule
\multicolumn{6}{@{}l}{\textit{Standard}} \\
Strong & \textbf{92.1} & 92.1 & 69.2 & 20.0 & 4.7 \\
Medium  & \textbf{84.9} & 77.3 & 72.9 & 27.2 & 7.8 \\
Weak  & \textbf{70.0} & 56.6 & 52.7 & 25.0 & 13.0 \\
\midrule
\multicolumn{6}{@{}l}{\textit{Real ambient light}} \\
Strong  & \textbf{89.9} & 89.3 & 67.7 & 22.2 & 4.7 \\
Medium  & \textbf{88.2} & 85.5 & 79.3 & 27.7 & 7.5 \\
Weak & \textbf{84.4} & 76.7 & 74.0 & 40.1 & 12.2 \\
\bottomrule
\end{tabular}
\label{tab:breakdown}
\end{table}

\begin{table}[t]
\centering
\small
\caption{Median lighting error in degrees on the 800 renders of \cref{tab:advice}. Guess is the error of guessing the same direction for every image, as in \cref{fig:backbones}.}
\begin{tabular}{@{}lccc@{}}
\toprule
Model & Azimuth & Elevation & Error \\
\midrule
Guess & 44.5 & 26.0 & 52.0 \\
\midrule
Qwen3.5-27B & 43.5 & 24.4 & 53.2 \\
Grok 4.7 & 19.6 & 20.8 & 34.0 \\
Gemini 3.1 Pro & 17.9 & 19.6 & 31.1 \\
Claude Fable 5.1 & 14.0 & 20.9 & 29.9 \\
GPT-6.1 Sol & 9.8 & 14.3 & 20.1 \\
\midrule
LightMatch & \textbf{2.6} & \textbf{2.1} & \textbf{3.7} \\
\bottomrule
\end{tabular}
\label{tab:vlm-error}
\end{table}

We also request the four photographers with consistent choices in \cref{sec:q} to rate placement on the fourth test set, shown in the Rating column. Each trial shows the photo before the move and the photo at the nearest of the 27 lamp positions to a hidden method's advice, rated from $-3$ to $+3$ by how much the lighting changes~\cite{international1996methods}. LightMatch scores 2.4 on average, the highest for all four photographers, with per-photographer ratings in \cref{tab:rating}, so its advice also improves the lighting on real photos.

\begin{table}[t]
\centering
\small
\caption{Photographer ratings on the fourth test set. Each value is the mean rating of one photographer on a scale from $-3$ to $+3$, where positive values mean the lighting improves after the move. Photographers P4 and P6 are excluded by the consistency check in \cref{sec:q}.}
\label{tab:rating}
\setlength{\tabcolsep}{4pt}
\begin{tabular}{@{}lccccc@{}}
\toprule
 & & Fixed & & Ambient- & \\
 & DECA & Rembrandt & DPR & unaware & LightMatch \\
\midrule
P1   & $-0.6$ & 1.2 & 1.5 & 1.9 & \textbf{2.3} \\
P2   & 0.4    & 1.4 & 2.1 & 1.6 & \textbf{2.2} \\
P3   & $-1.1$ & 0.8 & 1.9 & 2.3 & \textbf{2.5} \\
P5   & $-1.5$ & 1.0 & 1.7 & 2.2 & \textbf{2.6} \\
\midrule
Mean & $-0.7$ & 1.1 & 1.8 & 2.0 & \textbf{2.4} \\
\bottomrule
\end{tabular}
\end{table}

Moreover, to study the movable range, we test three ranges, azimuth $\pm15^\circ$, $\pm35^\circ$ and $\pm75^\circ$ with elevation $\pm10^\circ$, $\pm20^\circ$ and $\pm45^\circ$, on the first two test sets, whose ambient light comes from different sources, with the faces fixed, in \cref{fig:range} and \cref{tab:s-range}. As the range grows, LightMatch loses only 6.7\% of its Share on average while the other methods lose at least 32.4\%, so LightMatch is the most robust. 

\subsection{Ablation Study}
\label{sec:ablation}

To evaluate the contribution of each part of LightMatch, we conduct an ablation study, as shown in \cref{tab:ablation}. When the lamp-off photo is removed and the lamp is read from the lamp-on photo alone (labeled as w/o lamp-off), the Share drops by 7.4, 6.8 and 5.7 points on the three test sets, respectively, which indicates that the lamp-off photo helps separate the ambient light. When the real-face composites are removed from training and only the renders are kept (labeled as w/o real-face), the Share on real faces falls from 73.2\% to 30.6\%, which indicates that the real-face composites make up for the shortcomings of rendered faces.

The VLM-related comparisons use a variant of the first test set, with 800 new renders generated the same way but without ambient light, as shown in \cref{tab:advice}. When the VLM gives advice without reading the light (labeled as w/o read-out), the Share drops from 96.2\% to 90.5\%, which indicates that reading the light matters. Qwen3.5-27B reads the light best on renders at its middle layers 31 to 32, with an error of 13.0$^\circ$, as shown in \cref{fig:probe}, and w/ VLM read-out reads the light from these layers, but linear probes on these layers rise to 40$^\circ$ to 67$^\circ$ on real faces, above the 35$^\circ$ error of guessing the same direction for every image, as detailed in \cref{sec:s-models}. DINOv3 has the lowest azimuth and elevation errors among five frozen backbones on both renders and real faces, with 15.6$^\circ$ and 13.0$^\circ$ on real faces, as shown in \cref{fig:backbones}, so the perception head uses DINOv3. When the VLM reads the light (labeled as w/ VLM read-out) or gives the advice (labeled as w/ VLM advice), the Share is 94.8\% and 94.0\%, slightly below LightMatch, which indicates that DINOv3 with the solver is sufficient. With five VLMs such as GPT-6.1 Sol giving advice directly, the Share falls to 25.2\% or below, which indicates that the perception head with the solver is more reliable. To find the cause, we compare the light reported by these VLMs before their advice with the true light, as shown in \cref{tab:vlm-error}. Qwen3.5-27B has an Error of 53.2$^\circ$, close to the 52.0$^\circ$ of guessing the same direction for every image, and even GPT-6.1 Sol has 20.1$^\circ$, against 3.7$^\circ$ for LightMatch, which indicates that these VLMs already fail in reading the light.

\subsection{Case Study}

We compare the lamp moves of each method on concrete examples in \cref{fig:examples}. \Cref{fig:examples}(a) is from the Real ambient light test set. The best lamp position is at azimuth $-2^\circ$ and elevation $36^\circ$. DECA misses in azimuth by $-45^\circ$ (positive toward the right hand of the subject, negative toward the left hand) and in elevation by $+60^\circ$ (positive above the best position, negative below). Fixed Rembrandt misses in azimuth by $+45^\circ$. DPR misses in elevation by $+30^\circ$. Ambient-unaware also misses in azimuth by $+45^\circ$. LightMatch moves the lamp exactly to the best position. \Cref{fig:examples}(b) is a real photo from the Real capture test set. Each photo is taken after the lamp is moved to the position chosen by the method. The lamp starts at azimuth $+60^\circ$. DECA moves it to $-15^\circ$, Fixed Rembrandt moves it to $+45^\circ$, DPR moves it to $-60^\circ$, and Ambient-unaware moves it to $+45^\circ$, with ratings at least 0.7 lower than LightMatch. LightMatch moves the lamp to $0^\circ$ with a rating of 2.3. LightMatch mainly fails in the following cases. A large error in the ambient estimate, for example ambient light at azimuth $+19^\circ$ estimated at $+84^\circ$, moves the recommended lamp azimuth to $-13^\circ$. A lamp weaker than the ambient light, for example 0.26 times the ambient light, shifts the recommended lamp elevation by $+25^\circ$. A start already close to the best position shifts the recommended lamp azimuth by $+5^\circ$. More examples are in \cref{sec:s-more}.

\section{Conclusion}

Lighting is an important part of taking a good portrait. This paper answers where the lamp should go. We first measure good light with judgments that photographers confirm. Building on this measure, we propose LightMatch, which reads the light with DINOv3 features and isolates the lamp with an on and an off photo. In unseen environments, on real faces of unseen people, and on our real captures, LightMatch outperforms all compared methods on nearly every metric, including the main metric Share on all three test sets where it is measured. However, when the ambient estimate has a large error or when the lamp is weaker than the room light, its advice still drifts. In future work, we will optimize LightMatch, including advising on lamp intensity, moving to several lamps, and updating its advice during live streaming.

{
\small
\bibliographystyle{ieeenat_fullname}
\bibliography{main}

@misc{haven2025poly,
  title={{Poly Haven}: The Public {3D} Asset Library},
  author={{Poly Haven}},
  year={2025}
}

@inproceedings{aksoy2018dataset,
  title={A dataset of flash and ambient illumination pairs from the crowd},
  author={Aksoy, Yagiz and Kim, Changil and Kellnhofer, Petr and Paris, Sylvain and Elgharib, Mohamed and Pollefeys, Marc and Matusik, Wojciech},
  booktitle={Proceedings of the European Conference on Computer Vision (ECCV)},
  pages={634--649},
  year={2018}
}

@inproceedings{el2024probing,
  title={Probing the {3D} awareness of visual foundation models},
  author={El Banani, Mohamed and Raj, Amit and Maninis, Kevis-Kokitsi and Kar, Abhishek and Li, Yuanzhen and Rubinstein, Michael and Sun, Deqing and Guibas, Leonidas and Johnson, Justin and Jampani, Varun},
  booktitle={2024 IEEE/CVF Conference on Computer Vision and Pattern Recognition (CVPR)},
  pages={21795--21806},
  year={2024},
  organization={IEEE}
}

@article{bradley1952rank,
  title={Rank analysis of incomplete block designs: I. the method of paired comparisons},
  author={Bradley, Ralph Allan and Terry, Milton E},
  journal={Biometrika},
  volume={39},
  number={3/4},
  pages={324--345},
  year={1952},
  publisher={JSTOR}
}

@inproceedings{cao2026artimuse,
  title={{ArtiMuse}: Fine-grained image aesthetics assessment with joint scoring and expert-level understanding},
  author={Cao, Shuo and Ma, Nan and Li, Jiayang and Li, Xiaohui and Shao, Lihao and Zhu, Kaiwen and Zhou, Yu and Pu, Yuandong and Wu, Jiarui and Wang, Jiaquan and others},
  booktitle={Proceedings of the IEEE/CVF Conference on Computer Vision and Pattern Recognition},
  pages={15313--15322},
  year={2026}
}

@inproceedings{chaturvedi2025synthlight,
  title={{SynthLight}: Portrait relighting with diffusion model by learning to re-render synthetic faces},
  author={Chaturvedi, Sumit and Ren, Mengwei and Hold-Geoffroy, Yannick and Liu, Jingyuan and Dorsey, Julie and Shu, Zhixin},
  booktitle={2025 IEEE/CVF Conference on Computer Vision and Pattern Recognition (CVPR)},
  pages={369--379},
  year={2025},
  organization={IEEE}
}

@article{chinchuthakun2026diffusionlight,
  title={{DiffusionLight-Turbo}: Accelerated light probes for free via single-pass chrome ball inpainting},
  author={Chinchuthakun, Worameth and Phongthawee, Pakkapon and Raj, Amit and Jampani, Varun and Khungurn, Pramook and Suwajanakorn, Supasorn},
  journal={IEEE Transactions on Pattern Analysis and Machine Intelligence},
  year={2026},
  publisher={IEEE}
}

@inproceedings{dahou2025vision,
  title={Vision-language models can't see the obvious},
  author={Dahou, Yasser and Huynh, Ngoc Dung and Le-Khac, Phuc H and Para, Wamiq Reyaz and Singh, Ankit and Narayan, Sanath},
  booktitle={2025 IEEE/CVF International Conference on Computer Vision (ICCV)},
  pages={24159--24169},
  year={2025},
  organization={IEEE}
}

@article{feng2021learning,
  title={Learning an animatable detailed {3D} face model from in-the-wild images},
  author={Feng, Yao and Feng, Haiwen and Black, Michael J and Bolkart, Timo},
  journal={ACM Transactions on Graphics (ToG)},
  volume={40},
  number={4},
  pages={1--13},
  year={2021},
  publisher={ACM, New York, NY, USA}
}

@article{gardner2017learning,
  title={Learning to predict indoor illumination from a single image},
  author={Gardner, Marc-Andr{\'e} and Sunkavalli, Kalyan and Yumer, Ersin and Shen, Xiaohui and Gambaretto, Emiliano and Gagn{\'e}, Christian and Lalonde, Jean-Fran{\c{c}}ois},
  journal={arXiv preprint arXiv:1704.00090},
  year={2017}
}

@article{georghiades2001few,
  title={From few to many: Illumination cone models for face recognition under variable lighting and pose},
  author={Georghiades, Athinodoros S. and Belhumeur, Peter N. and Kriegman, David J.},
  journal={IEEE transactions on pattern analysis and machine intelligence},
  volume={23},
  number={6},
  pages={643--660},
  year={2001},
  publisher={IEEE}
}

@book{grey2014master,
  title={Master lighting guide for portrait photographers},
  author={Grey, Christopher},
  year={2014},
  publisher={Amherst Media}
}

@inproceedings{hold2017deep,
  title={Deep outdoor illumination estimation},
  author={Hold-Geoffroy, Yannick and Sunkavalli, Kalyan and Hadap, Sunil and Gambaretto, Emiliano and Lalonde, Jean-Fran{\c{c}}ois},
  booktitle={2017 IEEE Conference on Computer Vision and Pattern Recognition (CVPR)},
  pages={2373--2382},
  year={2017},
  organization={IEEE}
}

@book{international1996methods,
  title={Methods for subjective determination of transmission quality},
  author={{ITU-T}},
  note={Recommendation P.800},
  year={1996},
  publisher={International Telecommunication Union}
}

@article{jiang2026before,
  title={Before the Shutter: Aesthetic and Actionable Portrait Photography Planning in {3D} Scenes},
  author={Jiang, Ruixiang and Chen, Chang Wen},
  journal={arXiv preprint arXiv:2605.30318},
  year={2026}
}

@article{kamoi2024visonlyqa,
  title={{VisOnlyQA}: Large vision language models still struggle with visual perception of geometric information},
  author={Kamoi, Ryo and Zhang, Yusen and Das, Sarkar Snigdha Sarathi and Zhang, Ranran Haoran and Zhang, Rui},
  journal={arXiv preprint arXiv:2412.00947},
  year={2024}
}

@inproceedings{kim2024switchlight,
  title={{SwitchLight}: Co-design of physics-driven architecture and pre-training framework for human portrait relighting},
  author={Kim, Hoon and Jang, Minje and Yoon, Wonjun and Lee, Jisoo and Na, Donghyun and Woo, Sanghyun},
  booktitle={2024 IEEE/CVF Conference on Computer Vision and Pattern Recognition (CVPR)},
  pages={25096--25106},
  year={2024},
  organization={IEEE}
}

@article{lee2005acquiring,
  title={Acquiring linear subspaces for face recognition under variable lighting},
  author={Lee, Kuang-Chih and Ho, Jeffrey and Kriegman, David J},
  journal={IEEE Transactions on pattern analysis and machine intelligence},
  volume={27},
  number={5},
  pages={684--698},
  year={2005},
  publisher={IEEE}
}

@inproceedings{legendre2019deeplight,
  title={{DeepLight}: Learning illumination for unconstrained mobile mixed reality},
  author={LeGendre, Chloe and Ma, Wan-Chun and Fyffe, Graham and Flynn, John and Charbonnel, Laurent and Busch, Jay and Debevec, Paul},
  booktitle={2019 IEEE/CVF Conference on Computer Vision and Pattern Recognition (CVPR)},
  pages={5911--5921},
  year={2019},
  organization={IEEE}
}

@article{li2017learning,
  title={Learning a model of facial shape and expression from {4D} scans},
  author={Li, Tianye and Bolkart, Timo and Black, Michael J and Li, Hao and Romero, Javier},
  journal={ACM Trans. Graph.},
  volume={36},
  number={6},
  pages={194--1},
  year={2017}
}

@article{liang2026luxdit,
  title={{LuxDiT}: Lighting estimation with video diffusion transformer},
  author={Liang, Ruofan and He, Kai and Gojcic, Zan and Gilitschenski, Igor and Fidler, Sanja and Vijaykumar, Nandita and Wang, Zian},
  journal={Advances in Neural Information Processing Systems},
  volume={38},
  pages={5015--5046},
  year={2025}
}

@article{liu2023visual,
  title={Visual instruction tuning},
  author={Liu, Haotian and Li, Chunyuan and Wu, Qingyang and Lee, Yong Jae},
  journal={Advances in neural information processing systems},
  volume={36},
  pages={34892--34916},
  year={2023}
}

@inproceedings{ma2025hpsv3,
  title={{HPSv3}: Towards wide-spectrum human preference score},
  author={Ma, Yuhang and Wu, Xiaoshi and Sun, Keqiang and Li, Hongsheng},
  booktitle={2025 IEEE/CVF International Conference on Computer Vision (ICCV)},
  pages={15086--15095},
  year={2025},
  organization={IEEE}
}

@inproceedings{murray2012ava,
  title={{AVA}: A large-scale database for aesthetic visual analysis},
  author={Murray, Naila and Marchesotti, Luca and Perronnin, Florent},
  booktitle={2012 IEEE conference on computer vision and pattern recognition},
  pages={2408--2415},
  year={2012},
  organization={IEEE}
}

@article{pandey2021total,
  title={Total relighting: learning to relight portraits for background replacement},
  author={Pandey, Rohit and Orts-Escolano, Sergio and LeGendre, Chloe and H{\"a}ne, Christian and Bouaziz, Sofien and Rhemann, Christoph and Debevec, Paul E and Fanello, Sean Ryan},
  journal={ACM Trans. Graph.},
  volume={40},
  number={4},
  pages={43--1},
  year={2021}
}

@article{pellacini2007lighting,
  title={Lighting with paint},
  author={Pellacini, Fabio and Battaglia, Frank and Morley, R Keith and Finkelstein, Adam},
  journal={ACM Transactions on Graphics (TOG)},
  volume={26},
  number={2},
  pages={9--es},
  year={2007},
  publisher={ACM, New York, NY, USA}
}

@inproceedings{phongthawee2024diffusionlight,
  title={{DiffusionLight}: Light probes for free by painting a chrome ball},
  author={Phongthawee, Pakkapon and Chinchuthakun, Worameth and Sinsunthithet, Nontaphat and Jampani, Varun and Raj, Amit and Khungurn, Pramook and Suwajanakorn, Supasorn},
  booktitle={2024 IEEE/CVF Conference on Computer Vision and Pattern Recognition (CVPR)},
  pages={98--108},
  year={2024},
  organization={IEEE}
}

@article{radford2021learning,
  title={Learning transferable visual models from natural language supervision},
  author={Radford, Alec and Kim, Jong Wook and Hallacy, Chris and Ramesh, Aditya and Goh, Gabriel and Agarwal, Sandhini and Sastry, Girish and Askell, Amanda and Mishkin, Pamela and Clark, Jack and others},
  journal={arXiv preprint arXiv:2103.00020},
  year={2021}
}

@inproceedings{ramamoorthi2001efficient,
  title={An efficient representation for irradiance environment maps},
  author={Ramamoorthi, Ravi and Hanrahan, Pat},
  booktitle={Proceedings of the 28th annual conference on Computer graphics and interactive techniques},
  pages={497--500},
  year={2001}
}

@inproceedings{schoeneman1993painting,
  title={Painting with light},
  author={Schoeneman, Chris and Dorsey, Julie and Smits, Brian and Arvo, James and Greenberg, Donald},
  booktitle={Proceedings of the 20th annual conference on Computer graphics and interactive techniques},
  pages={143--146},
  year={1993}
}

@article{simeoni2025dinov3,
  title={{DINOv3}},
  author={Sim{\'e}oni, Oriane and Vo, Huy V and Seitzer, Maximilian and Baldassarre, Federico and Oquab, Maxime and Jose, Cijo and Khalidov, Vasil and Szafraniec, Marc and Yi, Seungeun and Ramamonjisoa, Micha{\"e}l and others},
  journal={arXiv preprint arXiv:2508.10104},
  year={2025}
}

@inproceedings{srikanth2014computational,
  title={Computational rim illumination with aerial robots},
  author={Srikanth, Manohar and Bala, Kavita and Durand, Fr{\'e}do},
  booktitle={Proceedings of the Workshop on Computational Aesthetics},
  pages={57--66},
  year={2014}
}

@inproceedings{zhou2019deep,
  title={Deep single-image portrait relighting},
  author={Zhou, Hao and Hadap, Sunil and Sunkavalli, Kalyan and Jacobs, David W},
  booktitle={Proceedings of the IEEE/CVF international conference on computer vision},
  pages={7194--7202},
  year={2019}
}

@article{talebi2018nima,
  title={{NIMA}: Neural image assessment},
  author={Talebi, Hossein and Milanfar, Peyman},
  journal={IEEE transactions on image processing},
  volume={27},
  number={8},
  pages={3998--4011},
  year={2018},
  publisher={IEEE}
}

@article{xu2023imagereward,
  title={{ImageReward}: Learning and evaluating human preferences for text-to-image generation},
  author={Xu, Jiazheng and Liu, Xiao and Wu, Yuchen and Tong, Yuxuan and Li, Qinkai and Ding, Ming and Tang, Jie and Dong, Yuxiao},
  journal={Advances in Neural Information Processing Systems},
  volume={36},
  pages={15903--15935},
  year={2023}
}

@inproceedings{zhang2025scaling,
  title={Scaling in-the-wild training for diffusion-based illumination harmonization and editing by imposing consistent light transport},
  author={Zhang, Lvmin and Rao, Anyi and Agrawala, Maneesh},
  booktitle={International Conference on Learning Representations},
  volume={2025},
  pages={10728--10745},
  year={2025}
}

@article{fei2023split,
  title={{SPLiT}: Single portrait lighting estimation via a tetrad of face intrinsics},
  author={Fei, Fan and Cheng, Yean and Zhu, Yongjie and Zheng, Qian and Li, Si and Pan, Gang and Shi, Boxin},
  journal={IEEE Transactions on Pattern Analysis and Machine Intelligence},
  volume={46},
  number={2},
  pages={1079--1092},
  year={2023},
  publisher={IEEE}
}

@inproceedings{bulat2017far,
  title={How far are we from solving the {2D} \& {3D} face alignment problem? (and a dataset of 230,000 {3D} facial landmarks)},
  author={Bulat, Adrian and Tzimiropoulos, Georgios},
  booktitle={2017 IEEE international conference on computer vision (ICCV)},
  pages={1021--1030},
  year={2017},
  organization={IEEE}
}

@inproceedings{chen2026polar,
  title={{POLAR}: A portrait {OLAT} dataset and generative framework for illumination-aware face modeling},
  author={Chen, Zhuo and Yang, Chengqun and Su, Zhuo and Lv, Zheng and Gao, Jingnan and Zhang, Xiaoyuan and Yang, Xiaokang and Yan, Yichao},
  booktitle={Proceedings of the IEEE/CVF Conference on Computer Vision and Pattern Recognition},
  pages={28871--28881},
  year={2026}
}

@article{wu2023yunet,
  title={{YuNet}: A tiny millisecond-level face detector},
  author={Wu, Wei and Peng, Hanyang and Yu, Shiqi},
  journal={Machine Intelligence Research},
  volume={20},
  number={5},
  pages={656--665},
  year={2023},
  publisher={Springer}
}

@article{sun2019single,
  title={Single image portrait relighting},
  author={Sun, Tiancheng and Barron, Jonathan T and Tsai, Yun-Ta and Xu, Zexiang and Yu, Xueming and Fyffe, Graham and Rhemann, Christoph and Busch, Jay and Debevec, Paul and Ramamoorthi, Ravi},
  journal={ACM Trans. Graph},
  volume={38},
  number={4},
  pages={1--12},
  year={2019}
}

@book{hunter2021light,
  title={Light—science \& magic: An introduction to photographic lighting},
  author={Hunter, Fil and Biver, Steven and Fuqua, Paul and Reid, Robin},
  year={2021},
  publisher={Routledge}
}
}

\FloatBarrier
\appendix

\section{Alignment of $Q$ with Professional Photographers}
\label{sec:s-photo-study}

This section details how $Q$ is checked against professional photographers in \cref{sec:q}, including the voting page, the voting process and the results.

\begin{figure}[!htb]
\centering
\includegraphics[width=\linewidth]{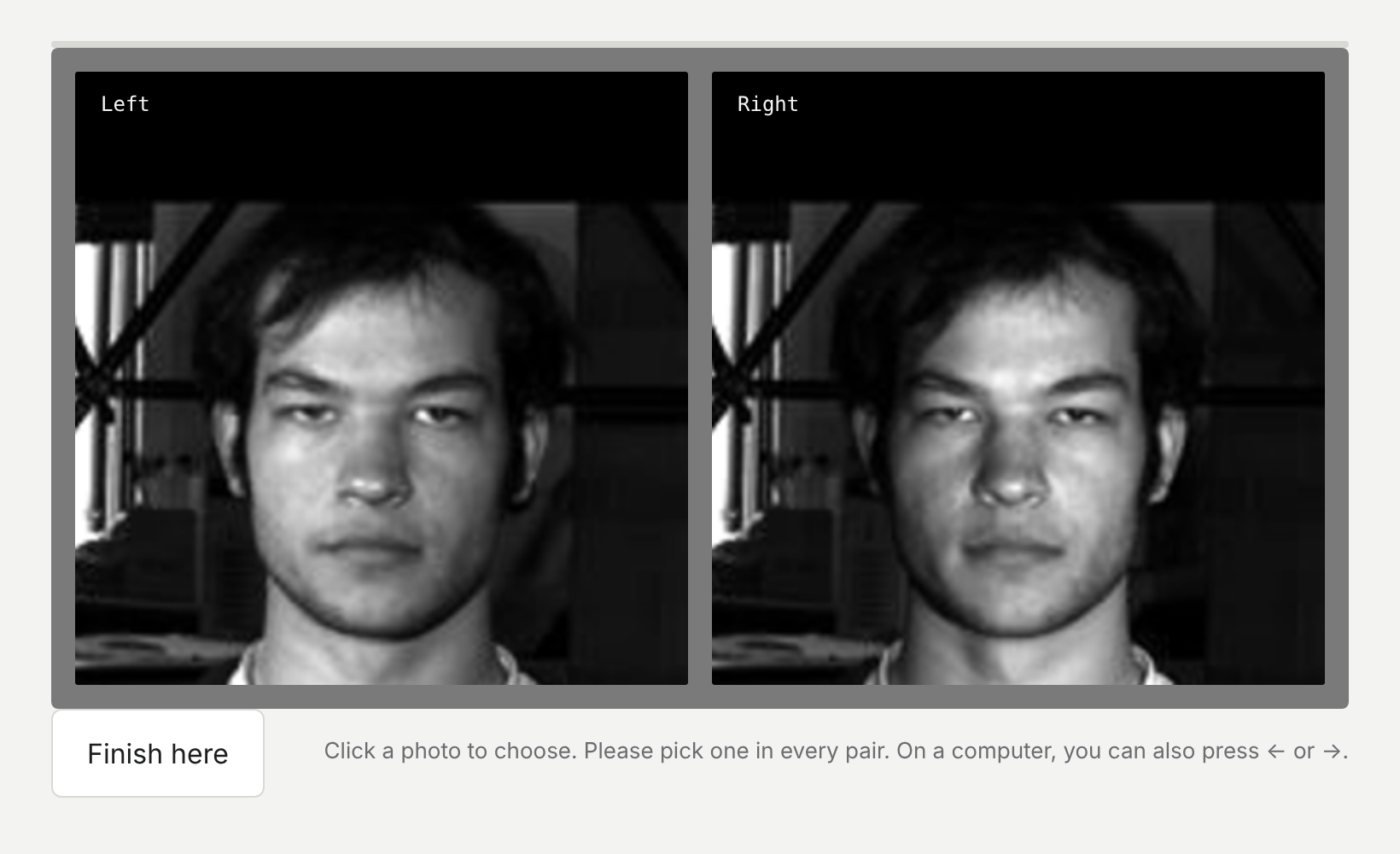}
\caption{\textbf{Voting page.} Both photos show the same subject, and the photographer clicks the one with better light.}
\label{fig:s-vote}
\end{figure}

\paragraph{Voting page.} \Cref{fig:s-vote} shows the page used by the photographers. Each trial shows two photos of the same Yale subject under different lamp positions and fill ratios, and the photographer clicks the one with better light. The real Yale faces~\cite{georghiades2001few,lee2005acquiring} are captured with a single lamp in a dark room, so the page adds the frontal image of the same subject, divided by 1.5, 3 or 6, as fill and equalizes the mean face brightness. Every trial requires a choice.

\begin{figure*}[t]
\centering
\includegraphics[width=\textwidth]{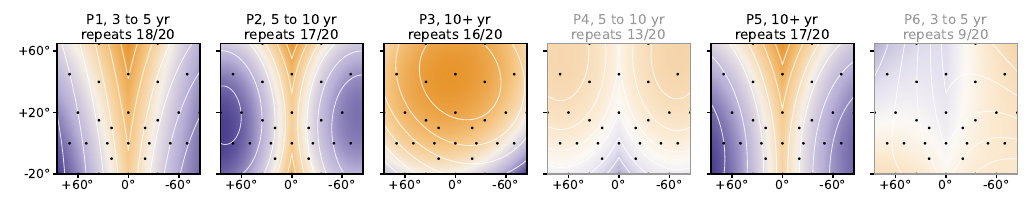}
\caption{$Q$ refitted on the votes of each photographer over the tested range for the fill at 1/3 of the frontal image, with purple to orange representing low to high $Q$ and dots at the 30 tested lamp positions. The title gives the experience and the number of repeats with the same choice. The two photographers in gray are discarded.}
\label{fig:s-photo}
\end{figure*}

\paragraph{Voting process.} The six photographers give 1,200 votes in total, with 20 repeats for each photographer. \Cref{tab:s-photo} lists the experience of each photographer and the number of repeats with the same choice. With the 80\% threshold in \cref{sec:q}, P1, P2, P3 and P5 are kept and P4 and P6 are discarded. \Cref{fig:s-photo} shows $Q$ refitted separately on the votes of each photographer.

\begin{table}[h]
\centering
\small
\caption{Experience and repeat consistency of the six photographers.}
\label{tab:s-photo}
\begin{tabular}{@{}lll@{}}
\toprule
Photographer & Experience & Consistent repeats \\
\midrule
P1 & 3 to 5 years & 18/20 \\
P2 & 5 to 10 years & 17/20 \\
P3 & more than 10 years & 16/20 \\
P4 & 5 to 10 years & 13/20 \\
P5 & more than 10 years & 17/20 \\
P6 & 3 to 5 years & 9/20 \\
\bottomrule
\end{tabular}

\end{table}

\paragraph{Results.} First, we compare how $Q$ and the photographers rank lamp positions. We refit $Q$ on all votes of the four photographers, and the refitted $Q$ has a Pearson correlation of 0.86 with $Q$ over the 30 lamp positions, with a 95\% confidence interval of $[0.79, 0.91]$, showing that $Q$ and the photographers largely favor the same positions. Second, we test whether $Q$ predicts the choices of the photographers. In pairs with the same fill ratio that differ only in lamp position, the agreement rate between $Q$ and the photographers is 64.7\%, which is the share of pairs where the photographers choose the image with the higher $Q$. As a reference, $Q$ refitted on the votes of the four photographers reaches an agreement rate of 72.8\% in 10-fold cross-validation. $Q$ is fitted without any of their votes, and its agreement rate is still close to this reference. When the $Q$ values of the two images differ by more than 1.5, the agreement rate rises to 84.4\%, showing that $Q$ and the photographers mainly disagree on pairs with similar $Q$.

\section{Datasets}
\label{sec:s-ambient}

This section adds details on the datasets in \cref{sec:setup}.

\paragraph{Rendered data.} The 300 CC0 Poly Haven~\cite{haven2025poly} panoramas used as ambient light are randomly rotated and mirrored during rendering. The second test set uses the panoramas of SPLiT as provided.

\paragraph{Real capture.} One person sits on a chair in one room. The nine lamp azimuths are spaced by $15^\circ$, and the three rings are set by the distance of the lamp. In daylight, the chair is placed at five positions and angles relative to the window, giving five conditions of window light, and each ring of nine lamp-on photos is followed by one lamp-off photo. At night, with only the ceiling light, the three rings are shot twice from the same seat and followed by one lamp-off photo. Faces are cropped to the head and shoulders. A lamp on the left of the person has a negative azimuth. \Cref{tab:s-real-new} lists the results by condition.

\begin{table}[t]
\centering
\small
\caption{Results on the fourth test set by capture condition. Lamp/ambient is estimated from the two photos. The last two columns report the median azimuth error in degrees.}
\label{tab:s-real-new}
\setlength{\tabcolsep}{2.8pt}
\resizebox{0.95\columnwidth}{!}{%
\begin{tabular}{@{}lccc@{}}
\toprule
Condition & Lamp/ambient & LightMatch & Ambient-unaware \\
\midrule
Night 2 & 4.2 & 6.1 & 8.1 \\
Night 1 & 2.8 & 9.9 & 10.5 \\
Daylight 4 & 3.3 & 15.2 & 19.6 \\
Daylight 3 & 7.8 & 15.3 & 17.2 \\
Daylight 2 & 1.0 & 37.1 & 63.9 \\
Daylight 1 & 0.6 & 42.4 & 60.5 \\
Daylight 5 & 0.8 & 68.2 & 97.7 \\
\midrule
All & - & 18.0 & 28.9 \\
Lamp/ambient $\geq2$ & - & 9.9 & 14.0 \\
\bottomrule
\end{tabular}%
}
\end{table}

\paragraph{Real faces from POLAR.} The third test set and the real-face training data are composited from POLAR~\cite{chen2026polar}. POLAR captures each subject under 157 single lights, 149 of them with a listed direction, from many cameras, and we use the ten subjects of its public preview with a neutral expression. Three choices are made from data. The camera is C12, whose all-lights image has the nose closest to the midpoint of the eyes. The azimuth origin of the light rig is fitted per subject from the left-right brightness asymmetry of the face across single-light images, which follows a sine of the azimuth. The left-right sign follows the light directions listed by POLAR relative to camera C12. A setting with ambient coefficients $(c_0,\mathbf{v})$ and a lamp of ratio $\rho$ at direction $\mathbf{u}$ is composited as $\sum_j w_j I_j$ over linearized single-light images $I_j$, where the ambient part sets $w_j=\Omega_j\max\!\big(0,\,c_0/(4\pi k_0)+3\,\mathbf{v}\!\cdot\!\mathbf{d}_j/(4\pi k_1)\big)$ with $\Omega_j$ the solid angle of light $j$ and $k_0,k_1$ the SH constants, and the lamp part spreads $c_0\rho/k_1$ over the three lights nearest $\mathbf{u}$. The lamp-off and lamp-on images share one exposure. Only the constant and first-order terms are used, matching what the method reads.

\paragraph{Training with real faces.} The ten subjects are split into two groups of five. One group is used for training and the other for testing, and then the two are swapped, so every subject is tested only by the model that did not see it. Real-face samples contribute loss only on the constant and first-order coefficients. 

\section{Implementation of Baselines and Variants}
\label{sec:s-yale}

This section gives the implementation of the baselines and ablation variants in \cref{sec:setup,sec:ablation}.

\begin{figure}[t]
\centering
\fbox{\parbox{0.95\linewidth}{\scriptsize\ttfamily
You are a portrait lighting expert. Look at this portrait and analyze its key light. Azimuth is the horizontal angle of the key light relative to the face in degrees (0 = frontal), elevation is the vertical angle (0 = eye level, positive = above), ratio is the key-to-fill light ratio. Report the current lighting, assess it, and recommend how to adjust it. Constraints: the key light can be moved horizontally by up to $\langle$L$\rangle$ degrees toward negative azimuth and up to $\langle$R$\rangle$ degrees toward positive azimuth; it can be lowered by up to $\langle$D$\rangle$ degrees and raised by up to $\langle$U$\rangle$ degrees; the person can turn their head by up to $\langle$T$\rangle$ degrees either way; the key-to-fill ratio can be scaled. Sign convention: positive azimuth means the key light comes from the left side of the image (as the viewer sees it), negative azimuth means the right side. Turning the head by a positive angle turns the face toward the left side of the image, which reduces the light's azimuth relative to the face by that angle.\\[3pt]
Answer in exactly this format and nothing else (numbers are degrees, no units):\\
Current lighting: azimuth $\langle$number$\rangle$, elevation $\langle$number$\rangle$, ratio $\langle$number$\rangle$.\\
Assessment: $\langle$one sentence$\rangle$.\\
Recommendation: move the light by $\langle$number$\rangle$ in azimuth and $\langle$number$\rangle$ in elevation, scale the ratio by $\langle$number$\rangle$, turn the head by $\langle$number$\rangle$.\\
Example:\\
Current lighting: azimuth -30, elevation +20, ratio 1.80.\\
Assessment: The key light is well off to one side.\\
Recommendation: move the light by +25 in azimuth and +10 in elevation, scale the ratio by 0.90, turn the head by +0.
}}
\caption{Prompt given to every VLM in \cref{tab:advice}. L, R, D, U and T are the movable range of each image, and each constraint changes to the corresponding wording when that move is not allowed.}
\label{fig:s-prompt}
\end{figure}

\paragraph{DPR and DECA.} The lighting estimates of DPR~\cite{zhou2019deep} and DECA~\cite{feng2021learning} are converted to our lighting representation. Both receive faces cropped by YuNet~\cite{wu2023yunet}. DECA uses a 1.25$\times$ face box at 224\,px with its own cropping turned off, as its FAN~\cite{bulat2017far} detector misses the face in many renders. Their SH axis conventions differ from those of our renderer, so we choose the axis permutation and signs on renders and fit a log-ratio map by least squares, with 2-fold cross-fitting that fits on one half of the images and tests on the other half. All lamp-placement experiments use this calibration.

\paragraph{Fixed Rembrandt and w/o lamp-off.} Fixed Rembrandt moves the lamp toward azimuth 45$^\circ$ and elevation 40$^\circ$, clipped to the movable range, and is given the true lamp position. The w/o lamp-off variant in the ablation study (\cref{sec:ablation}) uses only the lamp-on photo, with the mean ambient light of the training set.

\paragraph{VLM variants.} The w/ VLM read-out and w/ VLM advice variants in the ablation study (\cref{sec:ablation}) come from the same model. The first takes the output of the read-out, and the second takes the advice written by the model. The model is trained with LoRA on Qwen3.5-27B, and its read-out attends to the image tokens of layers 31 to 32. The training data have 4,000 renders and 840 Yale crops of the training subjects, giving 4,840 examples, and training runs for 3 epochs, or 909 steps, in about 17\,h on one A100. The w/o read-out variant uses the same training data with plain LoRA fine-tuning, without the read-out, so the model can state the light only in text. All models write advice with the prompt in \cref{fig:s-prompt}.

\section{Additional Results}
\label{sec:s-range}

This section gives additional data for the main results in \cref{sec:main}.

\begin{figure*}[t]
\centering
\includegraphics[width=\textwidth]{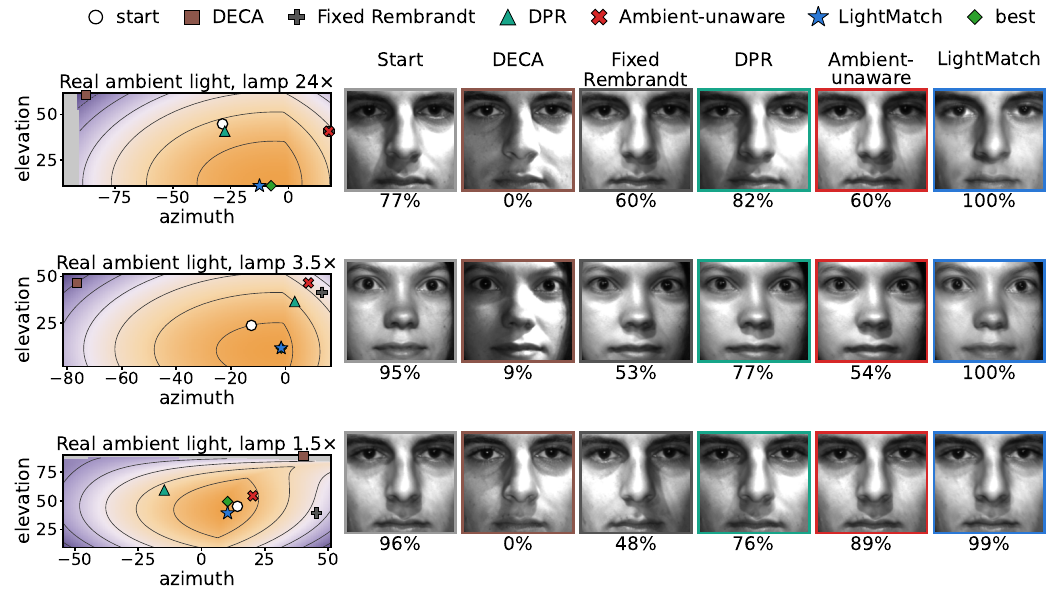}
\caption{Three more examples from the second test set. The map shows $Q$ of the combined light at each lamp position within the movable range, from purple (low) to orange (high), with gray outside the range covered by human votes. The markers show the start, the position chosen by each method, and the best position. The number below each face is the percentage of the best $Q$ within the movable range, measured from the lowest $Q$ in the range.}
\label{fig:s-more}
\end{figure*}

\begin{figure*}[t]
\centering
\includegraphics[width=\textwidth]{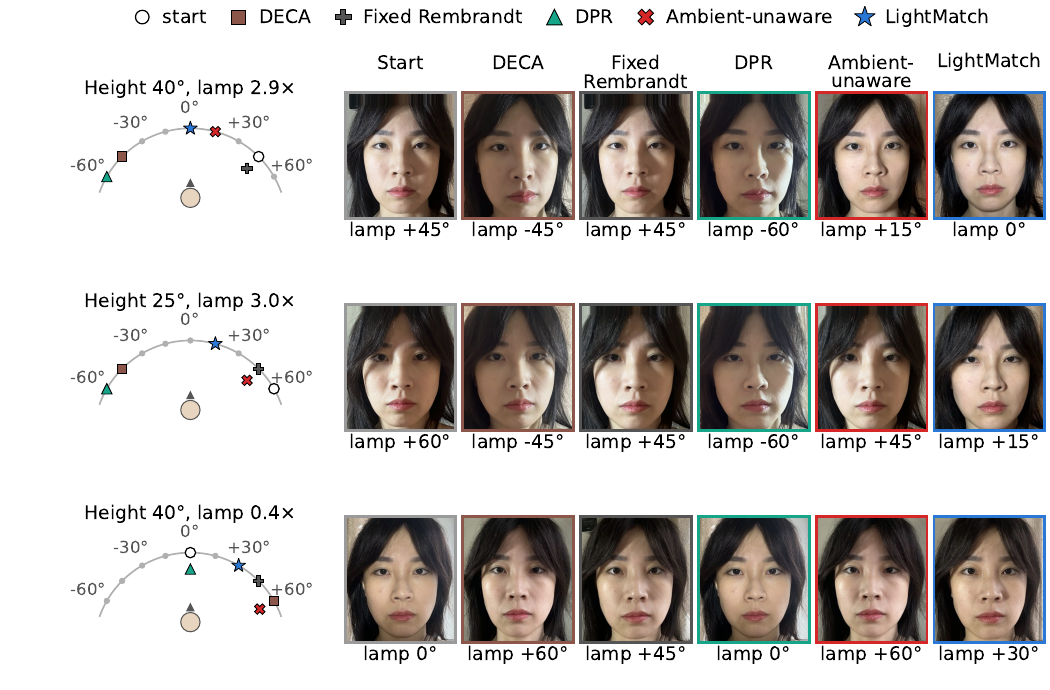}
\caption{Three real photos from the fourth test set. The title of each row gives the lamp height and the lamp strength relative to the ambient light. The diagram on the left of each row shows the nine lamp positions seen from above, with the start and the position chosen by each method, and the lamp position is below each photo.}
\label{fig:s-real}
\end{figure*}

\paragraph{Movable range.} \Cref{tab:s-range} gives the exact Share under each movable-range setting of \cref{fig:range}.

\begin{table*}[t]
\centering
\small
\caption{Share (\%) under different movable ranges of the lamp. Both enlarges the azimuth and elevation ranges together, written as azimuth/elevation. Azimuth only enlarges the azimuth range with elevation kept at $\pm10^\circ$. Elevation only enlarges the elevation range with azimuth kept at $\pm15^\circ$. Mixed enlarges one range to the middle setting and the other to the largest setting. All four settings start from $\pm15^\circ/\pm10^\circ$.}
\label{tab:s-range}
\resizebox{\textwidth}{!}{%
\begin{tabular}{llccccccccc}
\toprule
 & & \multicolumn{3}{c}{Both} & \multicolumn{2}{c}{Azimuth only} & \multicolumn{2}{c}{Elevation only} & \multicolumn{2}{c}{Mixed} \\
\cmidrule(lr){3-5}\cmidrule(lr){6-7}\cmidrule(lr){8-9}\cmidrule(lr){10-11}
Test set & Method & $\pm15^\circ/\pm10^\circ$ & $\pm35^\circ/\pm20^\circ$ & $\pm75^\circ/\pm45^\circ$ & $\pm35^\circ$ & $\pm75^\circ$ & $\pm20^\circ$ & $\pm45^\circ$ & $\pm35^\circ/\pm45^\circ$ & $\pm75^\circ/\pm20^\circ$ \\
\midrule
\multirow{5}{*}{Standard}
 & DECA            & 50.6 & 24.8 & $-$32.8 & 37.7 & $-$6.7 & 28.0 & $-$14.1 & $-$12.9 & $-$13.8 \\
 & Fixed Rembrandt & 44.6 & 28.1 & 2.7 & 24.3 & $-$3.1 & 45.4 & 42.4 & 27.7 & 1.3 \\
 & DPR             & 62.8 & 52.2 & 40.7 & 54.7 & 45.6 & 58.7 & 53.1 & 48.8 & 43.8 \\
 & Ambient-unaware & 84.0 & 74.1 & 55.0 & 77.1 & 64.8 & 76.5 & 64.0 & 65.0 & 62.6 \\
 & LightMatch      & \textbf{92.9} & \textbf{88.1} & \textbf{85.4} & \textbf{89.7} & \textbf{87.1} & \textbf{89.7} & \textbf{85.8} & \textbf{86.4} & \textbf{86.3} \\
\midrule
\multirow{5}{*}{Real ambient light}
 & DECA            & 39.7 & 16.3 & $-$31.0 & 29.1 & $-$16.6 & 25.3 & $-$9.2 & $-$8.8 & $-$20.4 \\
 & Fixed Rembrandt & 48.7 & 32.7 & 5.0 & 27.4 & $-$6.7 & 51.3 & 49.2 & 35.0 & 0.4 \\
 & DPR             & 68.9 & 56.4 & 42.8 & 59.3 & 49.0 & 63.2 & 54.9 & 49.7 & 46.9 \\
 & Ambient-unaware & 86.9 & 77.7 & 60.5 & 79.9 & 66.8 & 82.1 & 68.4 & 69.8 & 66.5 \\
 & LightMatch      & \textbf{93.0} & \textbf{90.8} & \textbf{88.0} & \textbf{91.2} & \textbf{89.1} & \textbf{92.6} & \textbf{90.3} & \textbf{89.7} & \textbf{88.8} \\
\bottomrule
\end{tabular}%
}

\end{table*}

\begin{table}[t]
\centering
\caption{Share (\%) computed with the $Q$ refitted on the votes of the four photographers. The lamp positions are those chosen by each method in \cref{tab:main}.}
\label{tab:pro_q}
\resizebox{\columnwidth}{!}{%
\begin{tabular}{lccc}
\toprule
Method & Standard & Real ambient light & Real faces \\
\midrule
DECA & 5.5 & 3.7 & 6.2 \\
Fixed Rembrandt & 29.7 & 28.2 & 23.4 \\
DPR & 60.3 & 55.1 & 31.5 \\
Ambient-unaware & 66.2 & 73.6 & 58.6 \\
LightMatch & \textbf{83.1} & \textbf{85.9} & \textbf{69.3} \\
\bottomrule
\end{tabular}%
}
\end{table}

\paragraph{Share under the photographers' $Q$.}
We keep the lamp position chosen by each method and recompute Share with the $Q$ refitted on the votes of the four photographers in \cref{sec:s-photo-study}. \Cref{tab:pro_q} gives the results. LightMatch leads on all three test sets, so its advantage also holds under the judgment of the photographers.

\paragraph{Single-lamp optimum.}
With a single lamp, $Q$ peaks near butterfly lighting at azimuth 0$^\circ$ and elevation 35$^\circ$ (\cref{sec:q}). To check whether this position stays best under ambient light, we move the lamp toward it, clipped to the movable range, and compare with LightMatch in \cref{tab:single_opt}. With ambient light, the single-lamp optimum falls short of LightMatch, as the best lamp position moves with the ambient light. On the fourth test set, LightMatch recommends all nine azimuths, and $0^\circ$, the azimuth of the single-lamp optimum, accounts for only 16 of the 189 pairs (8.5\%), as shown in \cref{tab:azimuth_dist}.

\begin{table}[t]
\centering
\caption{Share (\%) of moving the lamp toward the single-lamp optimum of $Q$.}
\label{tab:single_opt}
\resizebox{\columnwidth}{!}{%
\begin{tabular}{lccc}
\toprule
Method & Standard & Real ambient light & Real faces \\
\midrule	
Single-lamp optimum & 34.3 & 42.7 & 33.8 \\
LightMatch & \textbf{87.8} & \textbf{89.1} & \textbf{73.2} \\
\bottomrule
\end{tabular}%
}
\end{table}

\begin{table}[!t]
\centering
\small
\caption{Azimuth recommended by LightMatch on the fourth test set. Each value gives the number and percentage of the 189 pairs for which LightMatch recommends that azimuth.}
\label{tab:azimuth_dist}
\begin{tabular}{@{}rcc@{}}
\toprule
Azimuth & Pairs & Percentage (\%) \\
\midrule
$-60^\circ$ & 15 & 7.9 \\
$-45^\circ$ & 10 & 5.3 \\
$-30^\circ$ & 7 & 3.7 \\
$-15^\circ$ & 11 & 5.8 \\
$0^\circ$ & 16 & 8.5 \\
$+15^\circ$ & 15 & 7.9 \\
$+30^\circ$ & 34 & 18.0 \\
$+45^\circ$ & 28 & 14.8 \\
$+60^\circ$ & 53 & 28.0 \\
\bottomrule
\end{tabular}
\end{table}

\section{Details of VLM Probes}
\label{sec:s-models}

This section adds details on the VLM probes in the ablation study (\cref{sec:ablation}). The renders in this section are generated in the same way as in \cref{sec:ablation}, each from a randomly generated FLAME face, with a single lamp and without ambient light. The real faces are from Yale.

\paragraph{Probes in \cref{fig:probe}.} \Cref{fig:probe} shows the azimuth error of a linear probe on each layer of Qwen3.5-27B. For each layer, we average all image tokens into one vector and fit a ridge regression from this vector to the 9 SH coefficients in \cref{eq:sh}. The probes use 400 renders with 5-fold cross-validation.

\paragraph{VLM probes on real faces.} \Cref{sec:ablation} reports that the error of these probes rises to 40$^\circ$ to 67$^\circ$ on real faces. For this result, we train the same probes on 800 renders and test 12 layers on the real Yale faces. Among the 12 layers, the lowest azimuth error is 40$^\circ$ and the highest is 67$^\circ$. Guessing the same direction for every photo gives 35$^\circ$ on the same photos, and every layer is worse than this. The predicted elevation is also uncorrelated with the true elevation. This shows that the lighting features in the middle layers of the VLM fail to transfer from renders to real faces.

\section{Prompt of the VLMs in \texorpdfstring{\cref{tab:advice}}{the advice table}}
\label{sec:s-leak}

This section supplements the five columns from GPT-6.1 Sol to Qwen3.5-27B in \cref{tab:advice}. In these columns, the photo and a prompt are given directly to the original VLM, without a read-out or a solver, and the model gives the advice by itself. \Cref{fig:s-prompt} gives the full prompt.

\section{More Examples}
\label{sec:s-more}

\cref{fig:s-more} shows three more examples from the second test set. In the first row, the lamp is 24 times the ambient light, stronger than the example in \cref{fig:examples}. LightMatch reaches 100\% of the best $Q$, and the other methods range from 0\% to 82\%. In the last two rows, the lamp starts close to the best position. LightMatch stays at 99\% or more of the best $Q$ after the move, and every other method makes the light worse. \cref{fig:s-real} shows three more real photos from the fourth test set, covering two lamp heights of $40^\circ$ and $12^\circ$, with the lamp 2.3 to 2.9 times the ambient light.

\end{document}